\documentclass[sigconf,nonacm]{acmart}
\makeatletter
\renewcommand{\@mkauthors@iii}{%
  \gdef\@currentauthors{}%
  \def\@currentaffiliation{}%
  \def\@author##1{%
    \ifx\@currentauthors\@empty
      \gdef\@currentauthors{##1}%
    \else
      \g@addto@macro\@currentauthors{, ##1}%
    \fi}%
  \def\and{}%
  \def\affiliation##1##2{\def\@currentaffiliation{##2}}%
  \def\email##1##2{}%
  \addresses
  \global\setbox\mktitle@bx=\vbox{%
    \unvbox\mktitle@bx
    \centering
    {\normalfont\Large\bfseries\@currentauthors\par}%
    \vskip .2em
    {\normalfont\normalsize\@currentaffiliation\par}%
    \medskip}%
}
\makeatother
\usepackage{multirow}
\usepackage{algorithm}
\usepackage{algorithmic}
\usepackage{listings}
\usepackage{tcolorbox}
\tcbuselibrary{breakable,skins,listings}
\AtBeginDocument{%
  }

\setcopyright{none}

\begin{document}

\title[Beyond Score Prediction]{Beyond Score Prediction: LLM-Based Essay Scoring and Feedback Generation via Reinforcement Learning with Rubric Rewards}

\author{Xuefeng Jin}
\authornote{Contact: myjinxf@163.com}
\author{Jiashuo Zhang}
\author{Teng Cao}
\author{Bin Yang}
\affiliation{
  \institution{Alibaba Cloud Computing, Alibaba Group}
  \country{}
}
\renewcommand{\shortauthors}{Jin et al.}

\begin{abstract}
Large language models (LLMs) have been widely applied to automated essay scoring (AES) and automated feedback generation (AFG). However, existing studies rely primarily on prompt engineering or supervised fine-tuning, while systematic research on reinforcement learning (RL) post-training and automated evaluation of feedback quality remains limited. We propose RLAES, a unified LLM framework that jointly optimizes essay scoring and feedback generation through RL. To make feedback quality measurable, interpretable, and usable for training, we introduce Rubric-based Feedback Evaluation (RFE), an essay-grounded feedback evaluation framework comprising 166 fine-grained binary rubric items and an LLM-as-judge. Building on RFE, we propose Adaptive Gated Feedback Optimization (AGFO), which activates rubric-based feedback rewards on demand during RL, reducing evaluation overhead while improving feedback quality. We also propose Adjacent Contrastive Reasoning (ACR) to improve ordinal score calibration by explicitly contrasting adjacent score levels. Experimental results show that RFE rankings closely align with expert preferences. On the ASAP benchmark, RLAES achieves the best scoring performance among LLM-based methods (QWK = 0.803), while human pairwise evaluation favors its feedback over both the initialization and score-only RL.
Code and RFE resources are available at \url{https://github.com/hellomuyi/RLAES}.

\end{abstract}



\maketitle

\section{Introduction}
Automated Essay Scoring (AES) and Automated Feedback Generation (AFG) are core applications of artificial intelligence in education \cite{ABUDALFA2026101041}.
In Web-based learning environments, automated assessment can support learner evaluation and personalized learning \cite{liu2026eduembed}.
For writing-support services, AES and AFG use natural language processing, machine learning, and large language models (LLMs) to score student essays and generate constructive feedback.
Together, they can reduce educators' grading workload while providing learners with timely, personalized diagnostic feedback that helps them identify weaknesses and improve their writing over time \cite{Li_2026}.

Research on AES dates back to the 1960s \cite{Sun_2025}. Traditional AES approaches have evolved from regression models relying on hand-crafted features \cite{Choi2026} to deep learning paradigms that formulate AES as numerical classification or regression with automatic representation learning. This evolution has progressed from convolutional neural networks (CNNs) and recurrent neural networks (RNNs) \cite{taghipour-ng-2016-neural} to more advanced Transformer-based pretrained language models, most notably the BERT family \cite{Li_2025,xie-etal-2022-automated,chen-etal-2025-mixture-ordered}. 
Recent LLM-based approaches to AES can be broadly divided into prompt engineering and fine-tuning. Prompt engineering studies have extensively explored zero-shot \cite{Sessler_2025,su-etal-2025-essayjudge} and few-shot learning \cite{Huang_2026,mansour-etal-2024-large}, prompts augmented with scoring rubrics \cite{mansour-etal-2024-large}, chain-of-thought (CoT) reasoning \cite{Xiao_2025}, pairwise comparisons \cite{Choi2026,cai2025rankthenscore,shibata-miyamura-2025-lces}, and multi-agent frameworks \cite{Wang_2026}. Fine-tuning studies focus mainly on supervised fine-tuning (SFT) \cite{Xiao_2025,ormerod2024automated,Johnsi_2025}. These studies have shown that fine-tuned LLMs, even smaller open-source ones, substantially outperform their extremely large proprietary counterparts equipped with a range of enhanced prompting strategies.

Beyond score prediction, less attention has been paid to AFG, including the evaluation and optimization of feedback, despite the pedagogical importance of formative feedback \cite{Liu_2024}.
Small models that perform competitively on AES typically use BERT-based encoder-only architectures and cannot generate feedback without an additional decoder.
Existing LLM-based feedback generation methods either directly prompt an LLM \cite{stahl-etal-2024-exploring,Sasaki_2026,ormerod2024automated} or perform SFT on feedback generated by strong reasoning models \cite{Xiao_2025}.
CEAES \cite{li-pan-2025-ceaes} jointly optimizes scoring and feedback through bidirectional RL in a BART-based encoder--decoder model, with a dedicated regression head for scoring and reference feedback for supervised generation.
Feedback evaluation often relies on annotated datasets covering a limited set of categories \cite{Sasaki_2026,Liu_2024}. Human evaluation can assess richer LLM-generated feedback \cite{Xiao_2025,ormerod2024automated}, but is costly and difficult to scale for optimization.
\citet{li-pan-2025-ceaes} automate evaluation by measuring semantic similarity between generated feedback and scoring rubric descriptions associated with the gold score.
More recently, FeedEval trains separate evaluators for specificity, helpfulness, and validity to select feedback candidates, demonstrating alignment with expert judgments \cite{chu-etal-2026-feedeval}.

Despite recent advances in LLM-based AES and AFG, jointly improving scoring and feedback for Web-based writing-support services poses two challenges.
On the one hand, optimizing scoring accuracy alone may degrade feedback quality, motivating joint reinforcement learning (RL) post-training of decoder-only LLMs for AES and AFG.
We study the setting where the training corpus contains essay--score pairs but no reference feedback for individual training essays.
Feedback-generating SFT requires target feedback texts, whose construction entails additional generation or annotation and quality control.
Our approach optimizes sampled feedback and score tokens in one autoregressive response using score supervision and rubric rewards.

On the other hand, feedback optimization requires fine-grained evaluation signals grounded in the student's essay.
Similarity to high-level scoring rubric descriptions may miss essay-specific feedback errors \cite{li-pan-2025-ceaes}.
FeedEval's validity component evaluates entailment from score-specific rubric descriptions, without directly verifying feedback claims against the essay \cite{chu-etal-2026-feedeval}.
Moreover, its fine-tuned evaluators rely on substantial curated data, and adapting them to new evaluation criteria may require additional data and training.
These considerations motivate explicit, essay-grounded item judgments without task-specific evaluator fine-tuning.

To address these challenges, we propose RLAES, a unified LLM-based RL post-training framework for AES and AFG.
First, we introduce Rubric-based Feedback Evaluation (RFE), a prompt-specific, essay-grounded framework comprising 166 binary rubric items and an LLM-as-judge. RFE provides a fine-grained automated measure of feedback quality that can serve as an RL reward.
Building on RFE, we propose Adaptive Gated Feedback Optimization (AGFO), which activates rubric-based feedback rewards on demand during RL. 
AGFO reduces LLM-as-judge overhead while jointly optimizing scoring accuracy and feedback quality, thereby preventing the feedback degradation caused by score-only RL.
The resulting RLAES model achieves the best performance among LLM-based AES methods. We additionally investigate Adjacent Contrastive Reasoning (ACR) as an inference-time scoring strategy for general-purpose LLMs without AES-specific post-training.
Our main contributions are summarized as follows:
\begin{itemize}
	\item We investigate rubric-reward RL post-training for jointly optimizing AES and AFG with a decoder-only LLM. The resulting RLAES jointly optimizes scoring accuracy and feedback quality and achieves the best performance among LLM-based methods.
	\item We introduce RFE, an essay-grounded framework that provides interpretable, item-level feedback evaluation for AFG.
	\item We propose AGFO, which adaptively schedules feedback-reward evaluations to reduce LLM-as-judge overhead during joint RL optimization of scoring and feedback.
\end{itemize}

\section{Preliminaries}
\subsection{Task Definition}
Let $\mathcal{D}=\{(e_i,s_i)\}_{i=1}^{N}$ denote a dataset, where $e_i$ represents the $i$-th essay together with its scoring context, such as the writing prompt, source text, and scoring rubric guidelines, and $s_i$ is its corresponding holistic score. 
The objective is to train a model on $\mathcal{D}$ that generalizes well to unseen essays, enabling accurate score prediction for AES and reliable feedback generation for AFG.
We focus on prompt-specific AES and AFG, where the writing prompts used for training and testing are drawn from the same set.
We define feedback (or explanation) as all components of an AES system's output other than the score prediction. We structure the output such that the model first generates feedback and then predicts a score. In this setting, the feedback can be viewed as a CoT-like intermediate rationale before the final score.

\subsection{RL Post-Training}
Reinforcement learning with verifiable rewards (RLVR) relies on explicit, verifiable binary rules as reward signals, such as output correctness. DeepSeek-R1 \cite{Guo_2025} demonstrated its effectiveness in improving reasoning capabilities. Related algorithms such as Group Relative Policy Optimization (GRPO) \cite{shao2024deepseekmathpushinglimitsmathematical} have been widely applied to domains with checkable answers, including mathematics and code.

For difficult-to-verify domains such as essay feedback evaluation, however, feedback quality cannot be determined by a single binary rule and instead requires multidimensional assessment. Directly applying RLVR in such settings would yield an overly coarse reward signal. Conversely, reinforcement learning from human feedback (RLHF) can use human preferences to handle open-ended outputs, but preference judgments are highly subjective and may induce reward hacking.
Rubrics as Rewards (RaR) \cite{gunjal2025rubricsrewardsreinforcementlearning} decomposes evaluation criteria into individually assessable rubric items, bridging the gap between strictly verifiable rewards and subjective human preferences.
It provides structured rewards for open-ended tasks and is directly compatible with on-policy RL algorithms such as GRPO.

Specifically, given an input prompt $x$ and a model response $\hat{y}$, the rubric reward is defined as

\begin{equation}
	\label{eq:rubric-reward}
	r(x,\hat{y})
	=
	\frac{
		\sum_{j=1}^{k} w_j c_j(x,\hat{y})
	}{
		\sum_{j=1}^{k} w_j
	},
\end{equation}
where $k$ is the number of rubric items, $w_j > 0$ is the weight of item $j$, and $c_j:(x,\hat{y})\mapsto\{0,1\}$ indicates whether the response $\hat{y}$ to the model input $x$ satisfies that item. 
An LLM-as-judge assigns each indicator $c_j$ independently, and all items $\{(w_j,c_j)\}_{j=1}^{k}$ define the rubrics.

\section{Approach}
\subsection{Joint RL Post-Training for AES and AFG}
\label{sec:rlaes}
RLAES uses a decoder-only LLM to generate essay-specific feedback followed by a score in a single response. We jointly optimize these outputs with GRPO using a score reward and an essay-grounded feedback reward. Given a student essay and its context $e$, the gold score $s$, the model-predicted score $\hat{s}$, the model-generated feedback $\hat{f}$, and the feedback rubrics $\mathcal{R}_{\mathrm{fb}}$, the total reward $r_{\mathrm{total}}$ is defined as
\begin{equation}
	\label{eq:total-reward}
	r_{\mathrm{total}}
	= r_{\mathrm{score}}(s, \hat{s})
	+ \lambda_f \cdot r_{\mathrm{feedback}}
	(e, \mathcal{R}_{\mathrm{fb}}, \hat{f}),
\end{equation}
where $\lambda_f$ is a hyperparameter controlling the weight of the feedback reward. The score reward is defined as
\begin{equation}
	\label{eq:score-reward}
	r_{\mathrm{score}}(s, \hat{s})
	= -\frac{\left|s - \hat{s}\right|}{\alpha \, M},
\end{equation}
where $M=s_{\max}-s_{\min}$ is the prompt-specific score range and $\alpha$ is a scaling factor. RFE supplies $r_{\mathrm{feedback}}$ by evaluating $\hat{f}$ against $\mathcal{R}_{\mathrm{fb}}$.

Section~\ref{sec:rfe} defines the feedback evaluation framework, and Section~\ref{sec:agfo} describes how AGFO schedules reward evaluation during training. When the gate is closed, the feedback reward is set to zero.

Throughout the experiments, \textbf{RLAES} denotes the full model trained with AGFO, and \textbf{RLAES w/o AGFO} denotes its ablation trained using only the score reward.

\subsection{Rubric-Based Feedback Evaluation (RFE)}
\label{sec:rfe}
\begin{figure}
	\centering
	\includegraphics[width=1\columnwidth]{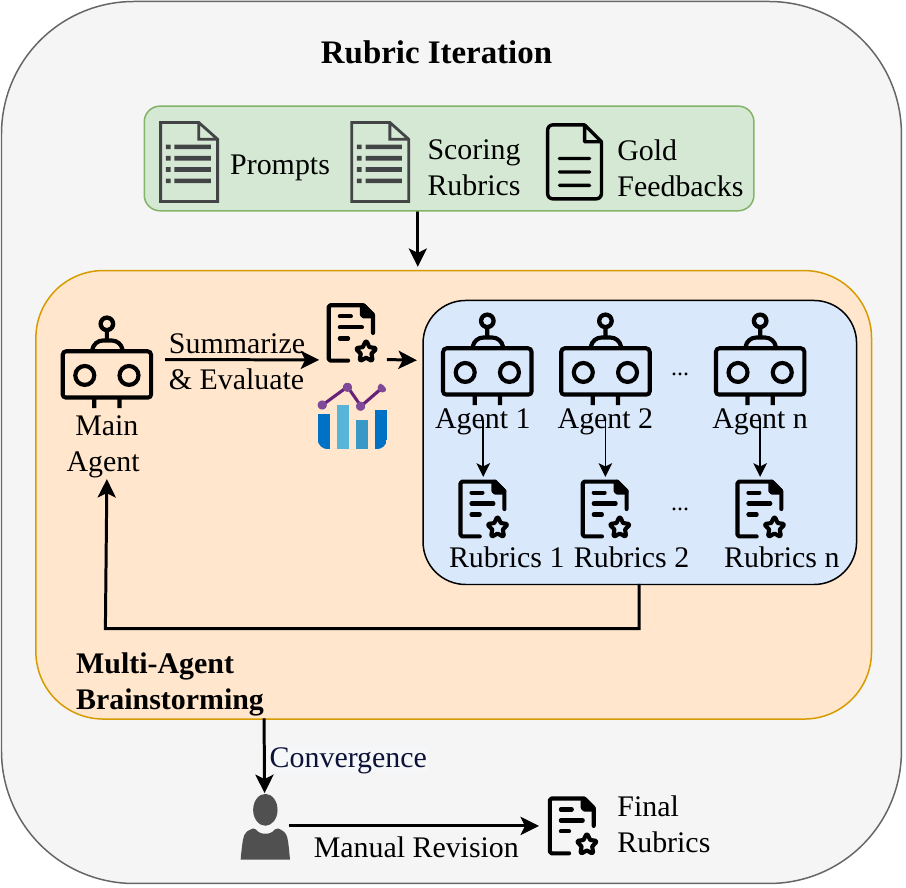} 
	\caption{Construction of the RFE rubrics.}
	\label{fig:rubrics_workflow}
\end{figure}

Inspired by Deep Research Bench II \cite{li-deepresearch-2026} and RaR \cite{gunjal2025rubricsrewardsreinforcementlearning}, we propose the Rubric-based Feedback Evaluation (RFE) framework, which supports automated feedback evaluation and converts feedback quality into a scalar reward for RL training. RFE comprises prompt-specific, checklist-style rubrics and an LLM-as-judge. Figure~\ref{fig:rubrics_workflow} illustrates the construction pipeline.

\paragraph{Gold Feedback Construction.}
We use Claude Opus 4.7 to extract handwritten essays and teacher feedback from official ASAP materials that are separate from the main ASAP dataset \cite{asap-aes}.
We remove feedback that is overly brief, merely repeats scoring rubric guidelines without essay-specific evidence, or omits too many trait domains, retaining 50 high-quality samples as pseudo-gold feedback.
These samples are then refined through LLM rewriting and expert revision, yielding 50 gold feedback samples.
These samples support rubric development and threshold calibration; they are not used as target responses for policy post-training.

\paragraph{Rubric Generation and Iteration.}
The Main Agent generates initial rubrics with manually adjusted top-level dimensions and fine-grained rubric items. It then evaluates the rubrics on the gold feedback and produces an analysis report. Five Rubric Refinement Agents---GPT-5.5, Gemini 3.1 Pro, Claude Opus 4.8, DeepSeek-V4-Pro, and GLM-5.1---independently propose revisions. The Main Agent aggregates these proposals and updates the rubrics until no further acceptable revision proposals are received.
After manual revision, the final rubrics, denoted by $\mathcal{R}_{\mathrm{fb}}$, contain 166 binary items across four dimensions: Coverage, Evidence, Faithfulness, and Safety.
Gold feedback and final rubrics were refined through collaborative expert discussion. Independent pre-discussion judgments were not recorded for these construction stages.

\paragraph{LLM-as-Judge Evaluation.}
Given an essay--feedback pair $(e_i,f_i)$, an LLM-as-judge independently determines whether each rubric item is satisfied. The RFE score is then computed using Equation~\ref{eq:rubric-reward}. Similar to Deep Research Bench II \cite{li-deepresearch-2026}, we assign equal weight to all items ($w_j=1$); thus, each dimension's contribution is proportional to its number of items.

\subsection{Adaptive Gated Feedback Optimization (AGFO)}
\label{sec:agfo}
Invoking an LLM-as-judge at every optimization step incurs prohibitive time and computational costs. We therefore propose Adaptive Gated Feedback Optimization (AGFO), which adaptively controls the activation of the feedback reward during training. 
Specifically, at training steps that are multiples of $n$, the system invokes the LLM-as-judge to compute the current model's feedback reward $r_{\mathrm{feedback}}$ on the batch and jointly optimizes scoring and feedback.
For $0 \leq \tau < 1$, if the batch mean feedback reward is below the predefined threshold $\tau$, the gate remains open, and the judge is queried and the resulting feedback reward is included in the total reward at every subsequent step until the mean reaches or exceeds $\tau$. Otherwise, the gate is closed between periodic checks, and the total reward contains only the score reward.
Pseudocode for AGFO is provided in Algorithm~\ref{alg:agfo}.

Notably, AGFO unifies two special cases via the threshold $\tau$. When $\tau = 0$, the feedback reward is activated only every $n$ steps, reducing AGFO to periodic feedback optimization. When $\tau = 1$, it is activated at every step, reducing AGFO to full feedback optimization.

\begin{algorithm}[t]
	\caption{Adaptive Gated Feedback Optimization (AGFO)}
	\label{alg:agfo}
	\begin{algorithmic}[1]
		\REQUIRE policy $\pi_\theta$; dataset $\mathcal{D}$; rubrics $\mathcal{R}_{\text{fb}}$;
		judge $\mathcal{J}$; period $n$; threshold $\tau$; weight $\lambda_f$; total steps $T$
		\ENSURE optimized policy $\pi_\theta$
		
		\STATE $\bar{r}_{\text{fb}} \gets 0$
		
		\FOR{$t \gets 1$ \TO $T$}
		\STATE Sample mini-batch $\mathcal{B} \subset \mathcal{D}$
		\STATE Generate responses $\{(\hat{f}_i,\hat{s}_i)\} \sim \pi_\theta(\cdot \mid e_i)$, $e_i \in \mathcal{B}$
		
		\IF{$t \bmod n \neq 0$ \AND $\tau < 1$ \AND $\bar{r}_{\text{fb}} \geq \tau$}
		\STATE $r^{\text{fb}}_i \gets 0,\ \forall i$ \COMMENT{gate closed}
		\ELSE
		\STATE $r^{\text{fb}}_i \gets r_{\text{feedback}}(e_i,\mathcal{R}_{\text{fb}},\hat{f}_i)$
		\COMMENT{gate open, query $\mathcal{J}$}
		\STATE Update $\bar{r}_{\text{fb}}$ with $\{r^{\text{fb}}_i\}$
		\ENDIF
		
		\FORALL{$i \in \mathcal{B}$}
		\STATE $r_i \gets r^{\text{score}}_i + \lambda_f\, r^{\text{fb}}_i$
		\ENDFOR
		
		\STATE Update $\pi_\theta$ via GRPO using $\{r_i\}$
		\ENDFOR
		
		\RETURN $\pi_\theta$
	\end{algorithmic}
\end{algorithm}

\subsection{Adjacent Contrastive Reasoning (ACR)}
\label{sec:acr}
Sections~\ref{sec:rlaes}--\ref{sec:agfo} describe the RLAES framework. As a separate inference-time study, we investigate ACR for general-purpose LLMs without AES-specific post-training. RLAES training and inference do not use ACR.

Automated essay scoring can be formulated as an ordinal classification problem: score levels are ordered labels rather than independent classes. 
The main difficulty is determining whether an essay belongs to a given score level or an adjacent lower or higher level.
Standard prompting methods ask an LLM to produce a holistic evaluation and score but do not explicitly require comparisons at score boundaries. 
This omission may make the model more prone to predicting a score level adjacent to the gold score---a phenomenon we call \textbf{adjacent confusion}.

\begin{figure}
	\centering
	\includegraphics[width=1\columnwidth]{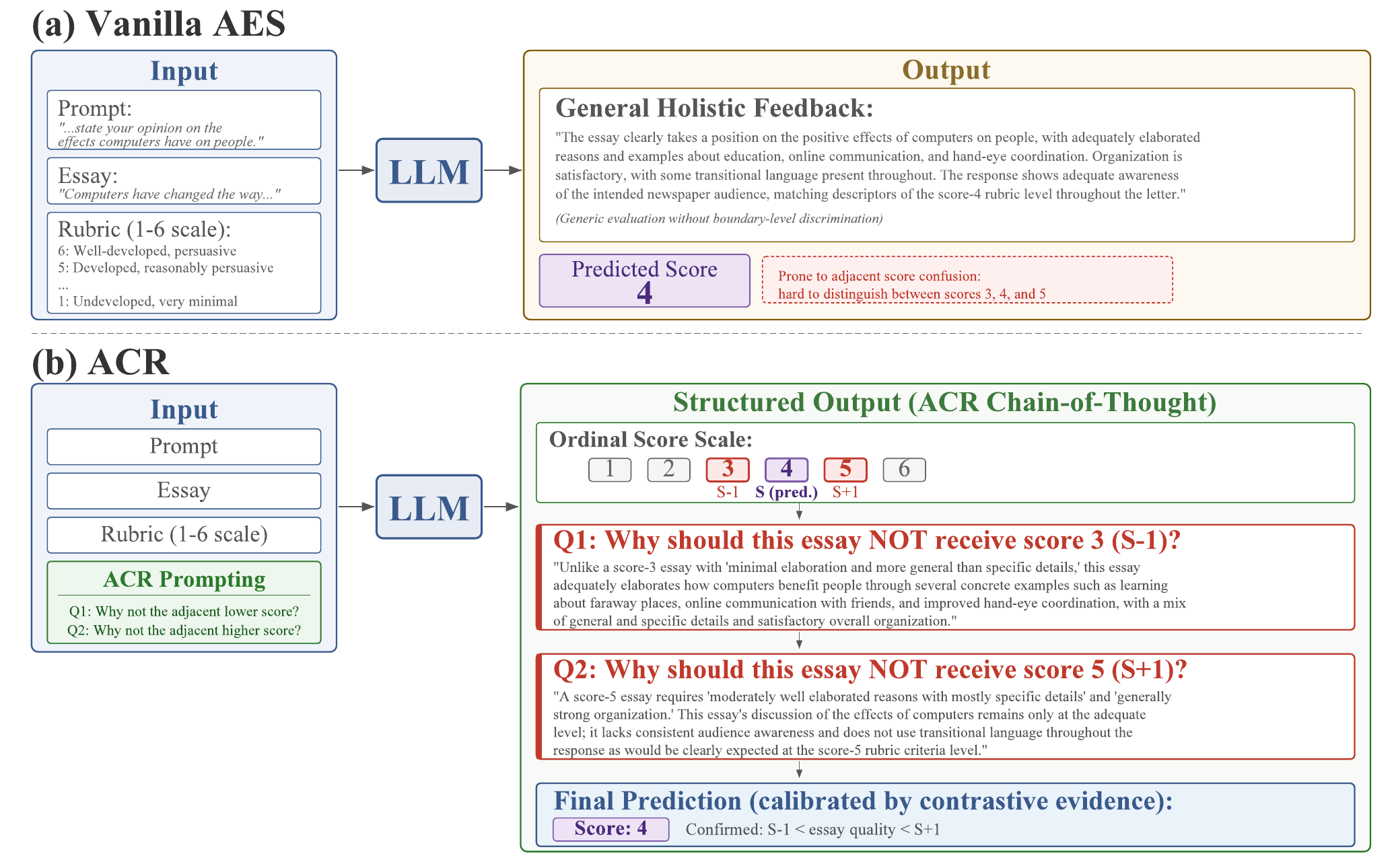} 
	\caption{ACR for ordinal score calibration.}
	\label{fig:ACR_pipeline}
\end{figure}

To alleviate this issue, we propose Adjacent Contrastive Reasoning (ACR), inspired by the anchoring-and-adjustment heuristic in human judgment \cite{tversky1974judgment}. As illustrated in Figure~\ref{fig:ACR_pipeline}, ACR treats the adjacent lower and higher scores around a predicted score as local hard-negative labels and requires the model to explicitly answer two contrastive questions before producing its final score: Why should the essay not receive the adjacent lower/higher score? This local counterfactual comparison encourages the model to identify evidence that distinguishes adjacent score levels instead of producing only a generic holistic assessment. ACR is designed to support scoring in three ways. First, it transforms scoring from direct score mapping into local boundary discrimination, thereby focusing the model's attention on fine-grained differences between adjacent scores. 
Second, treating adjacent levels as hard negatives directs attention toward more discriminative textual evidence.
Third, comparing both adjacent directions is intended to provide bidirectional calibration, helping reduce overestimation or underestimation.

ACR differs from existing few-shot prompting and pairwise-comparison methods, which introduce external reference essays and assist scoring by comparing the relative quality of a target essay against the reference.
In contrast, ACR requires no external reference samples and adds only local comparisons with adjacent score levels.
We also evaluate ACR-Lower, a one-sided variant that retains only the question of why the essay should not receive the adjacent lower score.
Extension to other ordinal tasks remains untested. Task-specific post-training may reduce the need for this guidance by learning score mappings from data.

\section{Experimental Setup}

\subsection{Dataset}
We evaluate on the Automated Student Assessment Prize (ASAP) dataset \cite{asap-aes}, a widely adopted benchmark for English AES. The dataset comprises 12,978 essays from three writing types and eight writing prompts (Table~\ref{tab:AES_statistics}). Following \citet{taghipour-ng-2016-neural}, we conduct five-fold cross-validation using their standardized train-dev-test splits. 
With one minor modification, we merge the training data from all eight prompts within the same fold and perform a single joint training run across prompts, rather than training a separate model for each prompt.
This design simplifies the experimental procedure, avoids model fragmentation, and enables parameter sharing across prompts.
For Web-based learning platforms supporting multiple writing assignments, a shared assessment model could reduce the need to train and maintain separate prompt-specific models.
We provide detailed results in Appendix~\ref{app:joint_training} showing that prompt joint training incurs negligible performance degradation across model families, with QWK decreasing by no more than 0.005.

\begin{table}
	\caption{Statistics of the ASAP dataset (Avg. Len.: average essay length; Arg.: argumentative; SD: source-dependent; Narr.: narrative).}
	\label{tab:AES_statistics}
	\centering
	 \setlength{\tabcolsep}{1mm}
	\begin{tabular}{llrrcr}
		\toprule
		\textbf{Prompt} & \textbf{Type} & \textbf{Grade} & \textbf{\# Essays} & \textbf{Avg. Len.} & \textbf{Range} \\
		\midrule
		1     & Arg.  & 8     & 1783   & 350  & 2--12   \\
		2     & Arg.  & 10    & 1800   & 350  & 1--6 \\
		3     & SD    & 10    & 1726   & 150  & 0--3   \\
		4     & SD    & 10    & 1772   & 150  & 0--3    \\
		5     & SD    & 8     & 1805   & 150  & 0--4   \\
		6     & SD    & 10    & 1800   & 150  & 0--4\\
		7     & Narr. & 7     & 1569   & 250  & 0--24\\
		8     & Narr. & 10    & 723    & 650  & 0--60\\
		\bottomrule
	\end{tabular}
\end{table}

\subsection{Evaluation Metrics}
We use quadratic weighted kappa (QWK), the most widely used metric in AES and the sole evaluation metric of the ASAP competition, to evaluate score prediction. QWK measures agreement between model predictions and human-assigned gold scores.

For AFG evaluation, we report human pairwise preferences, RFE scores, and feedback length on the same 240 essays, randomly sampled at 30 per prompt from the test set.
Human judgments are recorded as wins, ties, or losses for each model comparison.
RFE scores and average word and character counts provide complementary measures of rubric satisfaction and output length.

\subsection{Baselines and Prompting Strategies}
\label{sec:baselines}
Our primary scoring comparisons focus on decoder-only LLM-based AES methods, grouped into prompt engineering and post-training.
CEAES \cite{li-pan-2025-ceaes} is discussed as related work because its regression scoring head and reference-feedback supervision differ from our setting.

\paragraph{LLM Prompt Engineering.} These methods directly prompt LLMs to perform scoring without updating model parameters. We include GPT-3.5 with rubric-guided one-shot prompting \cite{mansour-etal-2024-large}, GPT-4 with rubric prompting \cite{li-pan-2025-ceaes}, GPT-4 with few-shot prompting \cite{Xiao_2025}, and Llama-3-8B-Instruct \cite{ormerod2024automated}. We also evaluate vanilla prompting with DeepSeek-V4-Pro \cite{deepseekai2026deepseekv4} and Qwen3.7-Max \cite{qwen37}, without AES-specific post-training.

\paragraph{ACR Prompting Study.}
For these two models, we compare vanilla prompting with ACR and ACR-Lower (Section~\ref{sec:acr}).

\paragraph{LLM Post-Training.} These methods adapt LLMs to AES by updating model parameters. 
Existing baselines in this category are primarily based on SFT, including fine-tuned GPT-3.5 \cite{DBLP:journals/corr/abs-2401-06431}, the Dual-Process Model \cite{Xiao_2025}, fine-tuned Llama-3.2-1B-Instruct \cite{Johnsi_2025}, fine-tuned Llama-3-8B-Instruct \cite{ormerod2024automated}, and RTS \cite{cai2025rankthenscore}.

For AFG, our controlled comparisons use the same initialization and no per-essay reference feedback targets for policy post-training.
They assess the effect of rubric-reward optimization; they do not establish superiority over feedback-supervised SFT, which requires an additional reference-feedback corpus.
We include GPT-5.6 Luna as a reference model in the automated feedback comparison.

\subsection{Implementation Details}
In our post-training experiments, we use Qwen3.5-9B \cite{qwen3.5} as the base model and adopt low-rank adaptation (LoRA) for parameter-efficient fine-tuning. For RLAES, we set the LoRA rank to 64 and the LoRA scaling parameter to 128. RLAES is trained for 25 epochs with a learning rate of $2.0 \times 10^{-5}$ and a warmup ratio of 0.1. We use a cosine learning-rate scheduler with a minimum learning-rate ratio of 0.1. The per-device training batch size is 8, with 4 gradient accumulation steps across 8 GPUs, giving an effective batch size of $8 \times 4 \times 8 = 256$ response sequences per optimizer step. For GRPO, we set the Kullback--Leibler (KL) regularization coefficient to 0. During rollout generation, we sample 8 responses for each query at a temperature of 0.9. All experiments are conducted on eight NVIDIA H20/A100 GPUs with DeepSpeed ZeRO-3.

The feedback reward weight $\lambda_f$ is set to 1, and the score-interval scaling factor $\alpha$ is set to 0.1. For the default AGFO setting, the gating period $n$ is set to 30, and the feedback reward threshold $\tau$ is set to 0.83 based on an offline evaluation of a small batch from the separate reference materials used for feedback extraction, outside the main ASAP dataset. We measure the model's current feedback quality, $r_{\mathrm{feedback}}$, by the current batch's mean RFE score.

We use two judges for offline RFE evaluation: GPT-6 Astra for the 240-essay feedback comparison and the case study, and GPT-5.5 \cite{openai2026gpt55} for the expert-alignment study and other RFE analyses. During RL training, DeepSeek-V4-Pro \cite{deepseekai2026deepseekv4} computes feedback rewards to limit the cost of repeated judge calls. These choices reflect evaluation and training costs; each reported comparison uses the same judge for all compared outputs. 

\section{Results and Analysis}
\subsection{Results on Score Prediction}

\begin{table*}[t]
	\caption{Scoring performance (QWK) across the eight ASAP prompts. Per-prompt values are rounded separately; AVG is computed from unrounded values and rounded only after averaging.}
	\label{tab:scoring_result}
	\centering
	\begin{tabular*}{\textwidth}{@{\extracolsep{\fill}}llccccccccc}
		\toprule
		\textbf{Type} & \textbf{Model} & \textbf{P1} & \textbf{P2} & \textbf{P3} & \textbf{P4} & \textbf{P5} & \textbf{P6} & \textbf{P7} & \textbf{P8} & \textbf{AVG} \\
		\midrule
		\multirow{10}{*}{\begin{tabular}{@{}l@{}}LLM \\Prompt\\Engineering\end{tabular}}
		& GPT-3.5 (rubric,1-shot) \shortcite{mansour-etal-2024-large}                              & 0.120 & 0.193 & 0.198 & 0.416 & 0.576 & 0.606 & 0.123 & 0.276 & 0.313 \\
		& GPT-4 (w/ rubric) \shortcite{li-pan-2025-ceaes}                      & 0.272 & 0.481 & 0.478 & 0.503 & 0.557 & 0.529 & 0.123 & 0.384 & 0.415 \\
		& GPT-4 (few-shot) \shortcite{Xiao_2025} & 0.280	&0.338	&0.331	&0.784	&0.623	&0.728	&0.257	&0.454	&0.474 \\
		&Llama-3-8B-Instruct \shortcite{ormerod2024automated}  	&0.255	&0.463	&0.432	&0.557	&0.653	&0.608	&0.283	&0.362	&0.452 \\
		& DeepSeek-V4-Pro (vanilla)             & 0.165 & 0.519 & 0.403 & 0.608 & 0.503 & 0.638 & 0.215 & 0.392 & 0.430 \\
		& \quad\textbf{+ ACR}             & 0.186 & 0.524 & 0.441 & 0.600 & 0.538 & 0.683 & 0.230 & 0.497 & 0.462 \\
		& \quad\textbf{+ ACR-Lower}       & 0.190 & 0.519 & 0.413 & 0.610 & 0.545 & 0.724 & 0.242 & 0.526 & 0.471 \\
		& Qwen3.7-Max (vanilla)                 & 0.231 & 0.517 & 0.396 & 0.483 & 0.545 & 0.650 & 0.184 & 0.565 & 0.446 \\
		& \quad\textbf{+ ACR}             & 0.248 & 0.500 & 0.487 & 0.594 & 0.593 & 0.668 & 0.217 & 0.478 & 0.473 \\
		& \quad\textbf{+ ACR-Lower}       & 0.275 & 0.525 & 0.499 & 0.633 & 0.660 & 0.716 & 0.230 & 0.508 & 0.506 \\
		\addlinespace
		\multirow{7}{*}{\begin{tabular}{@{}l@{}}LLM \\Post-Training\end{tabular}}
		& GPT-3.5 \shortcite{DBLP:journals/corr/abs-2401-06431}                   & 0.741 & 0.618 & 0.704 & \textbf{0.859} & 0.796 & \textbf{0.848} & 0.727 & 0.614 & 0.738 \\
		& Dual-Process Model \shortcite{Xiao_2025}                   & 0.761 & 0.652 & 0.724 & 0.809 & 0.812 & 0.776 & 0.707 & 0.489 & 0.716 \\
		& Llama-3.2-1B-Instruct \shortcite{Johnsi_2025}                & 0.711 & \textbf{0.778} & 0.682 & 0.723 & 0.769 & 0.692 & 0.776 & 0.631 & 0.720 \\
		& Llama-3-8B-Instruct \shortcite{ormerod2024automated}                 & 0.821 & 0.727 & 0.717 & 0.824 & 0.815 & 0.829 & 0.837 & 0.752 & 0.789 \\
		& RTS (Llama-3.1-8B) \shortcite{cai2025rankthenscore}                   & \textbf{0.840} & 0.712 & \textbf{0.752} & 0.844 & \textbf{0.831} & \textbf{0.848} & 0.830 & 0.732 & 0.798 \\
		& \textbf{SFTAES (ours)}                & 0.824 & 0.702 & 0.707 & 0.819 & 0.800 & 0.820 & 0.827 & 0.770 & 0.784 \\
		& \textbf{RLAES (ours)}            & 0.839 & 0.734 & 0.735 & 0.829 & 0.820 & 0.837 & \textbf{0.853} & \textbf{0.775} & \textbf{0.803} \\
		\bottomrule
	\end{tabular*}
\end{table*}

\paragraph{LLM Prompt Engineering and ACR} 
Table~\ref{tab:scoring_result} reports QWK for LLM prompt engineering and post-training methods. GPT-4 (w/ rubric) and GPT-4 (few-shot) achieve average QWK values of 0.415 and 0.474, respectively, substantially below post-trained LLMs. 
This gap highlights the difficulty of directly prompting LLMs for AES.

\begin{figure*}[t]
	\centering
	\includegraphics[width=0.88\textwidth]{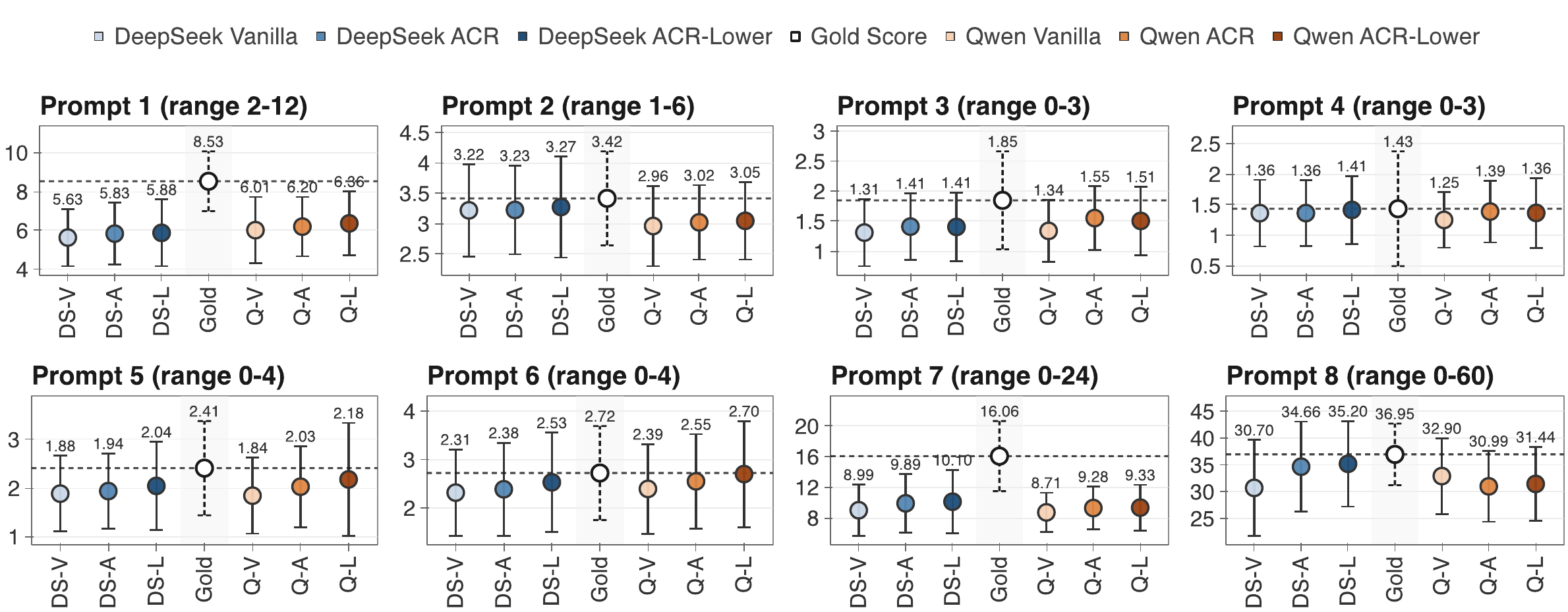} 
	\caption{Mean predicted scores and one-standard-deviation ranges under different ACR configurations. Dashed markers show prompt-level gold means.}
	\label{fig:ACR_res_A13_final}
\end{figure*}

For DeepSeek-V4-Pro, average QWK increases from 0.430 with vanilla prompting to 0.462 with ACR and 0.471 with ACR-Lower. For Qwen3.7-Max, the corresponding values are 0.446, 0.473, and 0.506. 
As shown in Figure~\ref{fig:ACR_res_A13_final}, both models systematically underestimate essay scores under vanilla prompting, whereas the one-sided ACR-Lower mitigates this bias by prompting them to identify the essay's strengths.
These results demonstrate improvements from ACR on the two evaluated models without AES-specific post-training.

\paragraph{Comparison with LLM Post-Training Methods.}
SFTAES, our SFT baseline that outputs only the predicted score, achieves an average QWK of 0.784, outperforming all prompt engineering baselines and several LLM post-training baselines.
RLAES achieves an average QWK of 0.803, exceeding RTS with Llama-3.1-8B (0.798), the strongest existing LLM post-training baseline in Table~\ref{tab:scoring_result}.
Compared with RTS, RLAES generates feedback and a score in a single response, requiring fewer model invocations and no reference-essay comparisons.
RTS performs up to 12 pairwise LLM inference runs involving reference essays in the Ranker stage to narrow the candidate score set, followed by one additional LLM inference run in the Scorer stage to predict the final score \cite{cai2025rankthenscore}.
Overall, RLAES achieves the highest average QWK among the LLM methods and is numerically close to the estimated human-like ceiling of 0.805 reported by \citet{uto2026has}.

\subsection{Results on Feedback Generation}
We randomly sample 30 essays from the fold 0 test set for each of the eight ASAP prompts, yielding 240 essays.
We compare the actual feedback generated by Step 0 (the Qwen3.5-9B initialization before RLAES post-training), RLAES w/o AGFO (trained solely with score rewards), and RLAES on this evaluation set.
For each essay, we conduct all three pairwise comparisons of the models' feedback.
To avoid circular evaluation based on RFE's own criteria, we adopt FeedEval's three dimensions---specificity, helpfulness, and validity---for these pairwise judgments \cite{chu-etal-2026-feedeval}. Two algorithm engineers familiar with ASAP receive training on these dimensions using a five-point Likert scale. During training, they rate feedback separately on each dimension to develop a shared understanding of the criteria and quality levels. They then divide the pairs between them, with each pair judged by one evaluator.
The presentation order of the two responses is randomized, and the evaluator selects the preferred feedback or a tie.
Appendix~\ref{app:inter_rater_reliability} reports an additional reliability assessment of the two evaluators.
We also report RFE scores and feedback lengths for these models and GPT-5.6 Luna on the same essays.

Table~\ref{tab:feedback_quality}(a) shows that human preferences favor RLAES over Step 0 and RLAES w/o AGFO, with win rates of 65.42\% and 95.83\%, respectively.
Step 0 is also preferred over RLAES w/o AGFO in 89.17\% of pairs.

The RFE scores in Table~\ref{tab:feedback_quality}(b) yield the same ordering: RLAES (0.679), Step 0 (0.618), and RLAES w/o AGFO (0.446).
The comparison with Step 0 shows higher human preference and RFE scores despite slightly shorter feedback on average: 366.51 vs.\ 381.98 words and 2427.12 vs.\ 2516.40 characters.
Given the complexity of feedback generation and evaluation, these aggregate statistics alone cannot rule out causal effects of feedback length on either evaluation measure.
Compared with score-only RL, RLAES achieves higher human preference and RFE scores with slightly shorter outputs on average (366.51 vs.\ 380.10 words and 2427.12 vs.\ 2742.58 characters).

GPT-5.6 Luna achieves the highest RFE score (0.798), with mean feedback lengths of 394.27 words and 2493.64 characters, and is not included in the human comparisons.
Its result serves as an automated reference at the observed output lengths, rather than evidence of a quality advantage at matched lengths.
The ablation study below examines these feedback gains together with the effect of AGFO on scoring accuracy.

\begin{table*}[t]
	\caption{Feedback evaluation on 240 test essays (30 per prompt). (a) Win/tie/loss rates for the first model in each comparison, with 240 pairs per comparison; ``w/o AGFO'' abbreviates RLAES w/o AGFO. (b) Mean RFE scores (GPT-6 Astra judge) and feedback lengths in words and characters.}
	\label{tab:feedback_quality}
	\centering
	\begin{minipage}[t]{0.48\textwidth}
		\centering
		\textbf{(a) Human pairwise preferences (\%)}\par\smallskip
		\begin{tabular*}{\linewidth}{@{\extracolsep{\fill}}lrrr@{}}
			\toprule
			\textbf{Comparison} & \textbf{Win} & \textbf{Tie} & \textbf{Loss} \\
			\midrule
			RLAES vs.\ w/o AGFO & 95.83 & 1.67 & 2.50 \\
			RLAES vs.\ Step 0 & 65.42 & 12.92 & 21.67 \\
			Step 0 vs.\ w/o AGFO & 89.17 & 6.67 & 4.17 \\
			\bottomrule
		\end{tabular*}
	\end{minipage}\hfill
	\begin{minipage}[t]{0.49\textwidth}
		\centering
		\textbf{(b) RFE scores and feedback length}\par\smallskip
		\begin{tabular*}{\linewidth}{@{\extracolsep{\fill}}lrrr@{}}
			\toprule
			\textbf{Model} & \textbf{RFE} & \textbf{Words} & \textbf{Chars.} \\
			\midrule
			GPT-5.6 Luna & \textbf{0.798} & 394.27 & 2493.64 \\
			Step 0 & 0.618 & 381.98 & 2516.40 \\
			RLAES w/o AGFO & 0.446 & 380.10 & 2742.58 \\
			\textbf{RLAES} & 0.679 & 366.51 & 2427.12 \\
			\bottomrule
		\end{tabular*}
	\end{minipage}
\end{table*}

\subsection{Evaluation of the RFE Framework}
\paragraph{Evaluation Objective.}
We conduct a meta-evaluation to assess RFE's alignment with human feedback preferences and examine the additional value of explicit, essay-grounded rubric evaluation over direct LLM judgment.
A comparison with a semantic-similarity metric alone cannot establish whether fine-grained rubric items contribute beyond an LLM's general ability to assess feedback.
We therefore compare GPT-5.5-based RFE with direct pairwise judgments from the same LLM, without RFE's item-level checklist.
Holding the judge model fixed allows us to examine the contribution of RFE's structured evaluation beyond the capabilities of the underlying LLM.

\paragraph{Human Preference Benchmark.}
We use an independent set of 68 ASAP essays not used for rubric development and construct two feedback pairs per essay, yielding 136 pairs.
Feedback candidates are generated using GPT-5.6 Sol and GPT-5.6 Luna to obtain responses of varying quality.
To reduce length bias, we regenerate candidates until the length difference between the two responses in each pair is below 10\%.
The same two evaluators independently judge all pairs, achieving 85.3\% agreement before discussion.
They resolve disagreements through discussion to establish consensus expert preferences.
All 136 length-matched pairs are retained for evaluation.

\paragraph{Compared Methods.}
We compare three automated evaluation approaches against the same consensus preferences on the same feedback pairs.
The BERTScore-based rubric-similarity method of \citet{li-pan-2025-ceaes} ranks feedback by semantic similarity to scoring rubric descriptions.
Both LLM-based methods use GPT-5.5.
The direct pairwise baseline compares the two feedback candidates directly, without RFE's fine-grained rubric items.
RFE instead evaluates each candidate against prompt-specific, essay-grounded binary rubric items and ranks the candidates by their aggregate RFE scores.
This same-judge baseline tests the additional value of RFE's evaluation framework more directly than a comparison with semantic similarity alone.
FeedEval~\cite{chu-etal-2026-feedeval} is also relevant, but relies on separately fine-tuned, dimension-specific evaluators; we could not access its trained checkpoints for a direct comparison.

\paragraph{Results and Interpretation.}
Table~\ref{tab:rfe_expert_agreement} shows that RFE achieves the highest observed agreement with consensus expert preferences, matching 125 of 136 pairs (91.9\%), compared with 107 pairs (78.7\%) for direct pairwise judging and 74 pairs (54.4\%) for BERTScore.
RFE exceeds the direct LLM baseline by 13.2 percentage points and BERTScore by 37.5 percentage points.
These results support the alignment of the evaluated RFE configuration with expert preferences on essays held out from rubric development.
Because both LLM-based methods use GPT-5.5 on the same feedback pairs, the higher agreement cannot be explained by the use of a different judge model.
This same-judge comparison provides empirical support for the additional contribution of RFE's explicit, essay-grounded rubric evaluation over direct LLM judgment in this setting.

\begin{table}[t]
	\caption{Agreement with consensus expert preferences on the same 136 length-matched feedback pairs. Both RFE and the direct pairwise baseline use GPT-5.5.}
	\label{tab:rfe_expert_agreement}
	\centering
	\setlength{\tabcolsep}{1mm}
	\begin{tabular*}{\columnwidth}{@{\extracolsep{\fill}}lr@{}}
		\toprule
		\textbf{Method} & \textbf{Agreement} \\
		\midrule
		BERTScore & 74/136 (54.4\%) \\
		Direct pairwise judge (GPT-5.5) & 107/136 (78.7\%) \\
		\textbf{RFE (GPT-5.5)} & \textbf{125/136 (91.9\%)} \\
		\bottomrule
	\end{tabular*}
\end{table}

\subsection{Ablation Study}
\paragraph{Effect of AGFO}
We compare RLAES with RLAES w/o AGFO to assess the contribution of feedback optimization.
Table~\ref{tab:agfo_ablation_qwk} combines five-fold QWK with RFE scores from the 240-essay feedback evaluation set.
Removing AGFO changes average QWK from 0.803 to 0.802 and reduces the RFE score from 0.679 to 0.446.
Human evaluators prefer the full model in 95.83\% of pairs (Table~\ref{tab:feedback_quality}).
These results support AGFO's benefit for the feedback produced under our generation settings while maintaining comparable scoring performance.

\begin{table*}[t]
	\caption{AGFO ablation on scoring and feedback quality. RFE scores are evaluated by GPT-6 Astra on the 240 test essays.}
	\label{tab:agfo_ablation_qwk}
	\centering
	\begin{tabular*}{\textwidth}{@{\extracolsep{\fill}}lcccccccccc@{}}
		\toprule
		\multirow{2}{*}{\textbf{Model}} & \multicolumn{9}{c}{\textbf{QWK (five-fold)}} & \multirow{2}{*}{\textbf{RFE}} \\
		\cmidrule(lr){2-10}
		& \textbf{P1} & \textbf{P2} & \textbf{P3} & \textbf{P4} & \textbf{P5} & \textbf{P6} & \textbf{P7} & \textbf{P8} & \textbf{AVG} & \\
		\midrule
		RLAES w/o AGFO & \textbf{0.840} & \textbf{0.739} & 0.732 & \textbf{0.829} & \textbf{0.821} & \textbf{0.839} & 0.846 & 0.770 & 0.802 & 0.446 \\
		\textbf{RLAES} & 0.839 & 0.734 & \textbf{0.735} & \textbf{0.829} & 0.820 & 0.837 & \textbf{0.853} & \textbf{0.775} & \textbf{0.803} & \textbf{0.679} \\
		\bottomrule
	\end{tabular*}
\end{table*}

\paragraph{Feedback Quality under Similar Activation Budgets.}
We compare the default adaptive setting $\tau=0.83$ ($n=30$) with periodic feedback optimization using $\tau=0$ ($n=21$).
The periodic and adaptive settings activate feedback rewards on 289 and 290 of 6075 steps, respectively, using the same batch size and rollout settings.
GPT-5.5 evaluates Prompt 1 test feedback from the validation-selected checkpoint for each setting.
Rubric-item win rate is the fraction of all 21 items with a higher mean pass rate than the other setting.
As shown in Table~\ref{tab:agfo_threshold}, $\tau=0.83$ achieves an RFE score of 0.867 versus 0.842 for $\tau=0$ and wins on 13 items (61.9\%), with four ties and four losses.
This comparison supports the benefit of adaptively allocating feedback-reward evaluations under similar activation budgets; broader threshold and period sensitivity remains to be studied.

\begin{table}[t]
	\caption{Feedback quality and training cost under different AGFO settings. The activation budgets differ by one step. Feedback metrics use GPT-5.5 as judge. Active steps count training steps with the feedback reward enabled. Counts and times cover all 6075 training steps. $^{*}$Full-activation time estimated from 72 consecutive feedback-active steps.}
	\label{tab:agfo_threshold}
	\centering
	\setlength{\tabcolsep}{2pt}
	\begin{tabular*}{\columnwidth}{@{\extracolsep{\fill}}lrrrr@{}}
		\toprule
		\textbf{Setting} & \textbf{RFE} & \shortstack{\textbf{Item win}\\\textbf{rate (\%)}} & \shortstack{\textbf{Active}\\\textbf{steps}} & \shortstack{\textbf{Time}\\\textbf{(h)}} \\
		\midrule
			$\tau=0$ ($n=21$) & 0.842 & 19.0 & 289 & 159 \\
		$\tau=0.83$ ($n=30$) & \textbf{0.867} & \textbf{61.9} & 290 & 156 \\
		\addlinespace
		$\tau=1$ & --- & --- & 6075 & $959^{*}$ \\
		\bottomrule
	\end{tabular*}
\end{table}

\begin{figure}[t]
	\centering
	\includegraphics[width=1\columnwidth]{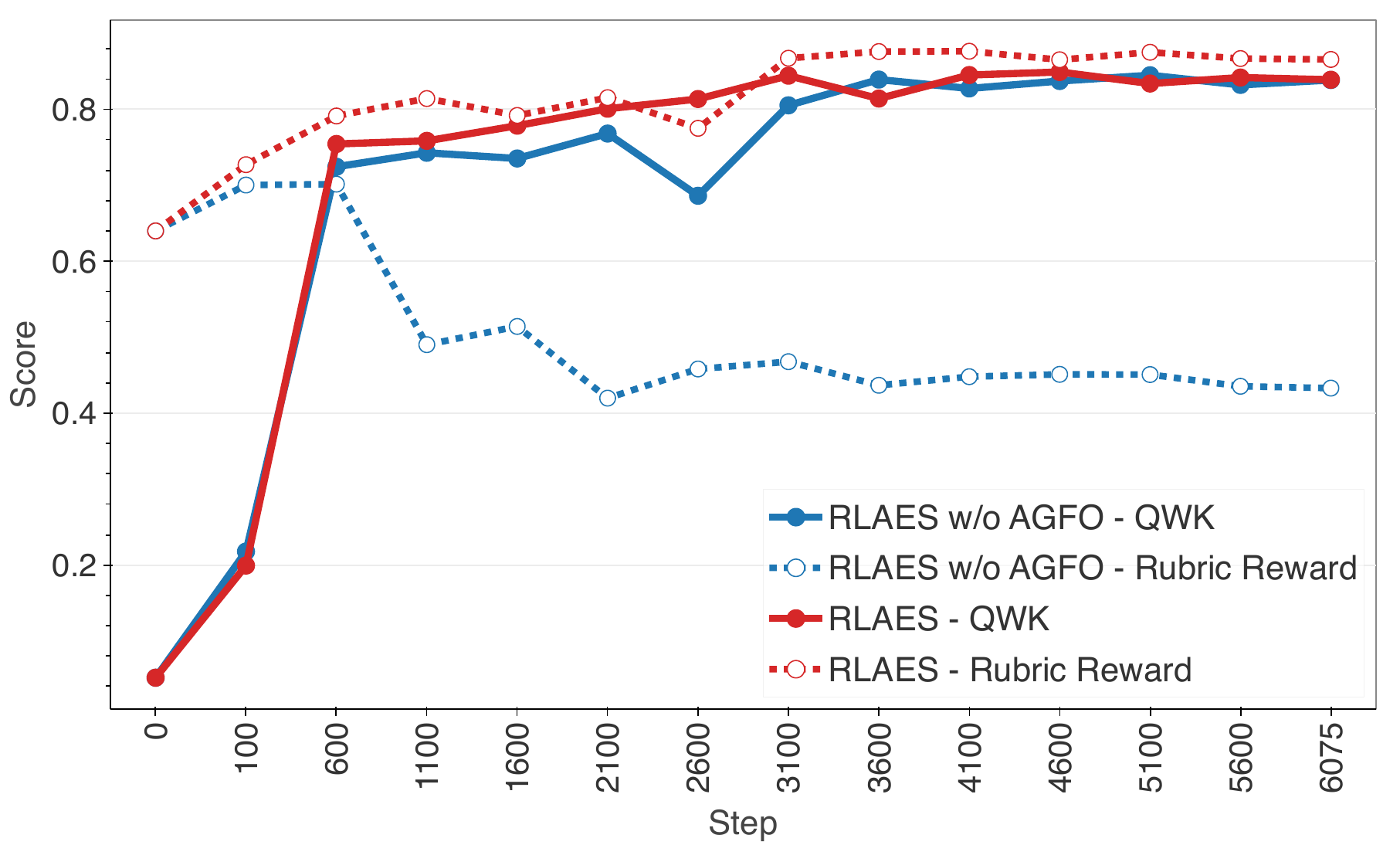} 
	\caption{QWK and rubric reward during RLAES w/o AGFO and RLAES training.}
	\label{fig:training_dynamics}
\end{figure}

\paragraph{Training Dynamics of Feedback Quality.}
Figure~\ref{fig:training_dynamics} tracks QWK and feedback quality (rubric reward, i.e., RFE score) on the Prompt 1 test set at selected training checkpoints, using GPT-5.5 as the RFE judge.
For RLAES w/o AGFO, QWK rises from 0.052 to above 0.83, whereas rubric reward briefly increases from 0.640 to 0.702 before falling to 0.433. 
Thus, improvements in scoring accuracy do not necessarily translate into better feedback. 
A plausible explanation is that the model initially uses feedback as an intermediate reasoning path for score prediction, temporarily improving feedback quality.
Once it learns to predict scores directly from input features, feedback no longer contributes to score optimization and, without direct optimization, begins to deteriorate.
By contrast, at the final checkpoint, RLAES achieves an RFE score of 0.866 versus 0.433 for RLAES w/o AGFO while maintaining comparable QWK, indicating that AGFO prevents feedback collapse without sacrificing scoring performance.
A finer-grained analysis shows higher final pass rates for RLAES across all four rubric dimensions than for RLAES w/o AGFO, with the largest improvement from initialization in Coverage.
Appendix~\ref{app:fine_grained_dynamics} provides dimension-level trajectories and item-level results, including remaining difficulties in quotation fidelity and accurate error correction.

\paragraph{Training Efficiency.}
The periodic ($\tau=0$, $n=21$) and adaptive ($\tau=0.83$, $n=30$) settings activate feedback optimization on 289 and 290 of 6075 training steps, respectively (both 4.8\%; Table~\ref{tab:agfo_threshold}). Their measured training times are 159 and 156 hours, respectively, showing comparable wall-clock cost under similar activation budgets. We estimate full-activation training time by multiplying the mean measured wall-clock time of 72 consecutive feedback-active steps by 6075, yielding approximately 959 hours for $\tau=1$. Relative to this extrapolated baseline, the measured adaptive run takes 83.7\% less time. This estimate assumes that the measured per-step cost is representative of the full run; a complete full-activation run was not conducted.

\section{Conclusion}
We present RLAES, an RL framework for jointly optimizing essay scoring and feedback generation. We introduce RFE for fine-grained feedback evaluation, yielding scores usable as RL rewards, and demonstrate that its rankings closely align with expert preferences.
Building on RFE, AGFO adaptively activates feedback rewards during RL, improving feedback quality over score-only RL while using feedback rewards on a small fraction of training steps. Its measured training time is lower than the extrapolated cost of full feedback optimization.
Experiments show that RLAES achieves the highest QWK among the LLM-based methods, while human pairwise preferences and RFE scores support its feedback gains over both the initial base model and score-only RL.
As an additional prompting study, ACR improves scoring performance on both evaluated general-purpose LLMs without AES-specific post-training.

RLAES remains limited by its dependence on LLM judges, whose differing preferences can cause reward mismatch, particularly for complex scoring rubrics. Expert-alignment results validate the GPT-5.5-based evaluation setting; agreement across the training and offline judges has not been measured. Evaluation is also restricted to prompt-specific English essays from ASAP. Future work should examine cross-prompt, cross-domain, and multilingual settings, as well as alternative strategies for joint scoring and feedback optimization.
Further evaluation in Web-based writing-support workflows should examine how learners use the generated feedback to revise their essays and how such feedback affects their writing proficiency.

\bibliographystyle{ACM-Reference-Format}
\bibliography{ref2027}



\appendix

\section*{Appendices}

\section{Joint Training Across Prompts}
\label{app:joint_training}
ASAP is commonly evaluated by performing five-fold cross-val\-i\-da\-tion independently for each of its eight prompts, requiring $8 \times 5 = 40$ training runs. Joint training reduces duplicated training and model maintenance across prompts.
Within each fold, we separately pool the training and validation splits across all eight prompts. We train one model on the pooled training set, select a single checkpoint using the pooled validation set, and evaluate that checkpoint separately on each prompt's test split. Because each input includes prompt-specific scoring guidelines, the model can handle different score ranges without normalization. This protocol reduces a complete evaluation from 40 to 5 training runs while preserving the original data partitions.

\paragraph{Evaluation of Joint Training Across Prompts.}
We evaluate prompt joint training with NPCR, SFTAES, and RLAES w/o AGFO, representing models spanning three method classes: the BERT-based AES model, LLM SFT, and LLM RL, respectively. NPCR is the most competitive open-source BERT-based baseline. The evaluation is conducted on fold 0.
As shown in Table~\ref{tab:ablation_joint_train}, joint training changes average QWK by $-0.0048$, $-0.0027$, and $-0.0041$, respectively. All decreases are below 0.005 for any of the three model classes, indicating only a negligible loss in scoring performance.

\begin{table*}[!t]
	\caption{Effect of prompt joint training on scoring performance (QWK). The RL experiments use RLAES w/o AGFO.}
	\label{tab:ablation_joint_train}
	\centering
	\setlength{\tabcolsep}{3pt}
	\begin{tabular*}{\textwidth}{@{\extracolsep{\fill}}lcccccccccc}
		\toprule
		\textbf{Model} & \textbf{P1} & \textbf{P2} & \textbf{P3} & \textbf{P4} & \textbf{P5} & \textbf{P6} & \textbf{P7} & \textbf{P8} & \textbf{AVG QWK} & $\boldsymbol{\Delta}$\textbf{QWK} \\
		\midrule
		NPCR w/o joint    & 0.8260 & 0.7170 & 0.7270 & 0.8650 & 0.8400 & 0.8340 & 0.7920 & 0.7410 & 0.7928 & -- \\
		NPCR     & 0.8334 & 0.7301 & 0.7328 & 0.8170 & 0.8267 & 0.8462 & 0.7880 & 0.7292 & 0.7879 & $-0.0048$ \\
		SFTAES w/o joint  & 0.8183 & 0.7006 & 0.7287 & 0.8303 & 0.8260 & 0.8466 & 0.8132 & 0.7401 & 0.7880 & -- \\
		SFTAES            & 0.8240 & 0.7044 & 0.7158 & 0.8244 & 0.8014 & 0.8319 & 0.8139 & 0.7663 & 0.7853 & $-0.0027$ \\
		\shortstack[l]{RLAES w/o AGFO\\(w/o joint)} & 0.8487 & 0.7392 & 0.7466 & 0.8509 & 0.8455 & 0.8498 & 0.8410 & 0.7452 & 0.8084 & -- \\
		RLAES w/o AGFO & 0.8331 & 0.7529 & 0.7620 & 0.8355 & 0.8332 & 0.8462 & 0.8333 & 0.7386 & 0.8043 & $-0.0041$ \\
		\bottomrule
	\end{tabular*}
\end{table*}

\section{Fine-grained Analysis of Feedback Training Dynamics}
\label{app:fine_grained_dynamics}
We analyze feedback on the Prompt 1 test set at 14 training checkpoints using GPT-5.5 as judge.
The 21 rubric items comprise Coverage (7), Evidence (4), Faithfulness (7), and Safety (3).
For each dimension, we average the pass rates of its constituent items.

\begin{figure}[!t]
	\centering
	\includegraphics[width=\columnwidth]{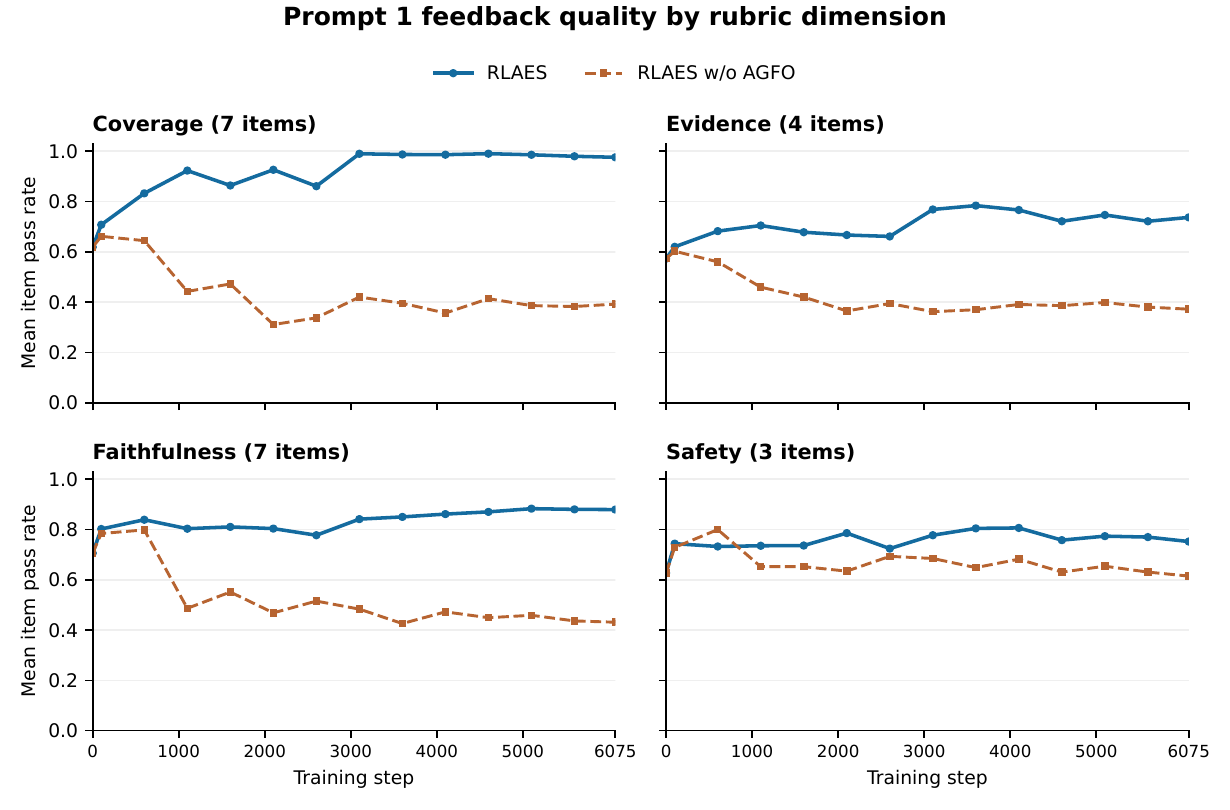}
	\caption{Mean rubric-item pass rates by dimension on the Prompt 1 test set during training, evaluated by GPT-5.5. Each dimension averages its constituent items with equal weights.}
	\label{fig:dimension_dynamics}
\end{figure}

\paragraph{Dimension-level Trends.}
Figure~\ref{fig:dimension_dynamics} shows that RLAES achieves higher final pass rates than RLAES w/o AGFO across all four dimensions.
From initialization to the final checkpoint, Coverage increases from 0.619 to 0.976, Evidence from 0.573 to 0.737, Faithfulness from 0.706 to 0.879, and Safety from 0.627 to 0.752.
The largest improvement occurs in Coverage, indicating more comprehensive feedback on the aspects specified by the rubric.

\begin{figure*}[!t]
	\centering
	\includegraphics[width=0.9\textwidth]{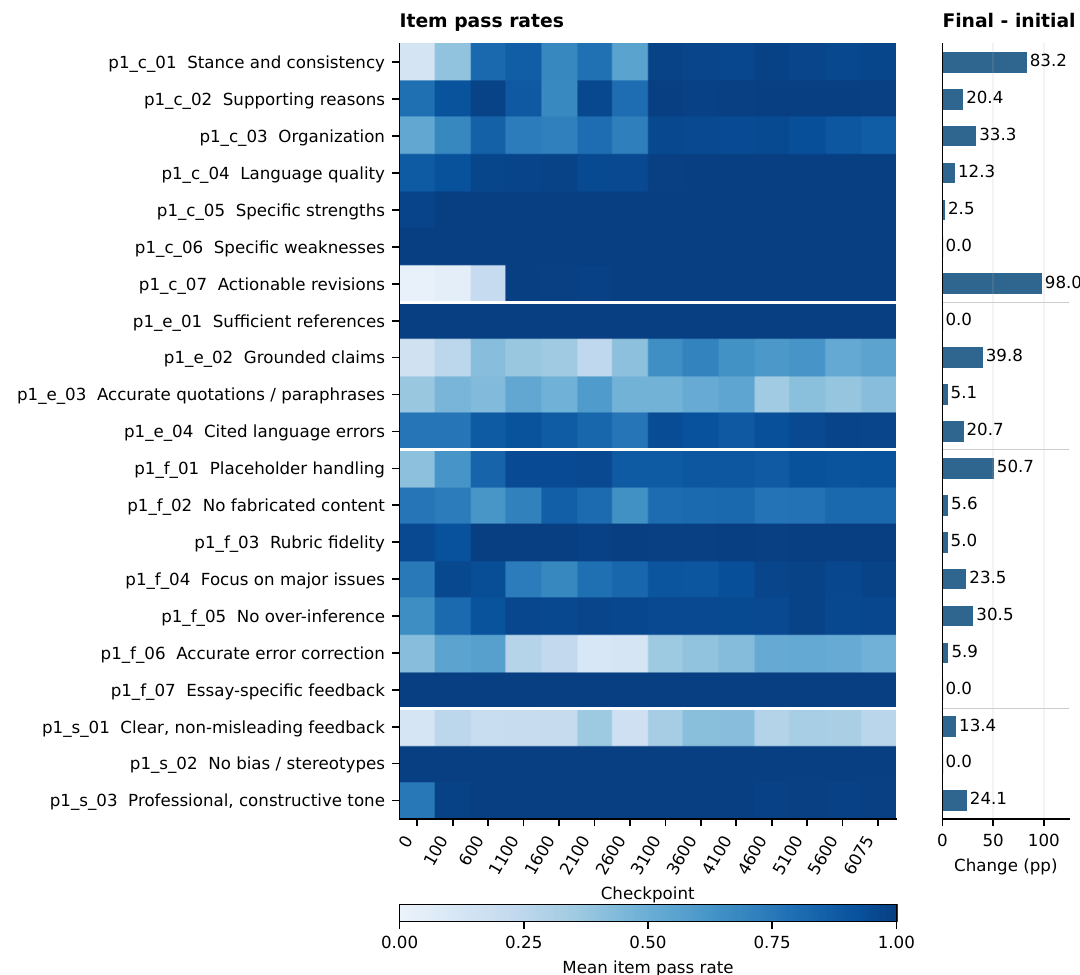}
	\caption{Item-level feedback quality for RLAES on Prompt 1, evaluated by GPT-5.5. Rows are grouped by Coverage, Evidence, Faithfulness, and Safety. Darker blue indicates a higher mean pass rate. Checkpoint columns are equally spaced; bars show final minus initial pass rates in percentage points (pp).}
	\label{fig:item_dynamics}
\end{figure*}

\paragraph{Item-level Trends.}
Figure~\ref{fig:item_dynamics} presents the trajectories of all 21 items for RLAES.
The largest gains include actionable revision suggestions (\texttt{p1\_c\_07}, 0.020 to 1.000) and accurate identification and evaluation of the student's stance (\texttt{p1\_c\_01}, 0.135 to 0.966).
Quotation fidelity (\texttt{p1\_e\_03}, 0.370 to 0.420) and accurate language-error correction (\texttt{p1\_f\_06}, 0.420 to 0.479) show smaller gains and remain challenging.
The trajectories also fluctuate across checkpoints, showing that the overall improvement combines substantial gains in some items with persistent weaknesses in others.

\newtcblisting[use counter=lstlisting]{paperprompt}[3][]{
  enhanced,
  breakable,
  listing only,
  listing engine=listings,
  title={Listing \thetcbcounter: #2},
  title after break={Listing \thetcbcounter: #2 (continued)},
  label={#3},
  colback=gray!10,
  colframe=gray!50!black,
  colbacktitle=gray!50!black,
  coltitle=white,
  fonttitle=\bfseries\small,
  boxrule=0.7pt,
  arc=1.5mm,
  left=2mm, right=2mm,
  top=1.5mm, bottom=1.5mm,
  before skip=8pt, after skip=8pt,
  lines before break=5,
  listing options={
    basicstyle=\rmfamily\small,
    columns=fullflexible,
    keepspaces=true,
    breaklines=true,
    breakatwhitespace=true,
    breakautoindent=true,
    breakindent=0pt,
    numbers=none,
    showstringspaces=false,
    frame=none,
    aboveskip=0pt, belowskip=0pt,
    moredelim={[is][\bfseries]{|}{|}}
  },
  #1
}



\section{Inter-Rater Reliability}
\label{app:inter_rater_reliability}

Both evaluators are algorithm engineers familiar with ASAP, hold master's degrees, and have passed the College English Test Band 6 (CET-6).

We randomly sampled 20 essays and their corresponding RLAES-generated feedback. Both evaluators independently scored each response using the RFE framework. The two-way random-effects, absolute-agreement, single-measure ICC (ICC(2,1)) was 0.774. In the independent pairwise preference evaluation of 136 feedback pairs, the same two evaluators achieved 85.3\% agreement before discussion. Together, these results suggest good agreement between the two evaluators in both absolute scoring and pairwise preference judgments.

\section{Case Study}
\label{app:case_study}
We compare four feedback responses for an ASAP Prompt 4 essay (essay ID 9669; gold score 3).

Table~\ref{tab:case_study} shows that RLAES satisfies 15 of 21 items, one fewer than GPT-5.6 Luna and more than Qwen3.5-9B (12) and RLAES w/o AGFO (10).
Across all four models, feedback lengths range from 2455 to 2658 characters, a maximum difference of 8.3\% relative to the shortest response.
RLAES distinguishes expressed information from implied meaning, gives concrete revision suggestions, and identifies grammatical errors with examples (\texttt{p4\_c\_01}, \texttt{p4\_c\_06}, and \texttt{p4\_e\_05}).
Qwen3.5-9B incorrectly attributes the mother's remark about the hibiscus surviving the cold to the grandmother (\texttt{p4\_e\_04}).
RLAES w/o AGFO provides no actionable revision advice (\texttt{p4\_c\_06}) and credits the student with unsupported analysis of family support (\texttt{p4\_f\_02}).
Complete responses appear in Table~\ref{tab:case_study_outputs}, and aggregate feedback comparisons are reported in Table~\ref{tab:feedback_quality}.

\begin{table}[H]
\caption{Feedback quality and length for the selected Prompt 4 case (essay ID 9669). RFE is the proportion of 21 rubric items satisfied according to GPT-6 Astra. Character counts include internal whitespace, and word counts use whitespace splitting.}
\label{tab:case_study}
\centering
\small
\setlength{\tabcolsep}{3pt}
\begin{tabular*}{\columnwidth}{@{\extracolsep{\fill}}lrrrr@{}}
\toprule
\textbf{Model} & \shortstack{\textbf{Rubric Items}\\\textbf{Passed}} & \shortstack{\textbf{RFE}\\\textbf{Score}} & \textbf{Chars.} & \textbf{Words} \\
\midrule
GPT-5.6 Luna & 16/21 & 0.762 & 2472 & 398 \\
Qwen3.5-9B & 12/21 & 0.571 & 2455 & 374 \\
RLAES w/o AGFO & 10/21 & 0.476 & 2515 & 355 \\
\textbf{RLAES (ours)} & 15/21 & 0.714 & 2658 & 405 \\
\bottomrule
\end{tabular*}
\end{table}

\begin{table*}[!t]
\centering
\caption{Complete feedback for the selected Prompt 4 case (essay ID 9669).}
\label{tab:case_study_outputs}
\footnotesize
\setlength{\tabcolsep}{3pt}
\renewcommand{\arraystretch}{0.95}
\begin{tabular}{@{}p{0.09\textwidth} p{0.895\textwidth}@{}}
\toprule
\textbf{Model} & \textbf{Feedback} \\
\midrule
GPT-5.6 Luna & Your response identifies several important elements of the ending: Saeng’s sadness, her memories of Vietnam, the hibiscus as a symbol of home, and her renewed determination to retake the test. Your reference to “Saeng broke down and wept” effectively shows how the greenhouse reconnects her with what she has lost. You also notice that the final paragraph changes the direction of the story toward hope. \\
& The main improvement is to explain more directly why the author ends the story with this paragraph, rather than mainly describing what Saeng plans to do. The ending brings together several earlier images: the failed driving test, the winter setting, the returning geese, and the hibiscus that will bloom in spring. These details create a pattern of renewal. Saeng’s plan to take the test again shows that her failure is temporary, just as winter is temporary and the hibiscus can survive the cold. The final paragraph therefore shows her developing ability to adapt while still maintaining a connection to Vietnam and her cultural memories. \\
& Your interpretation of the tone is partly accurate but needs more balance. You write that the story is “more of a depress[ing] mood,” but the conclusion is not hopeless. The sadness in the greenhouse and Saeng’s tears help make the hopeful ending meaningful. You could explain that the story moves from grief and homesickness toward resilience. Also, the source says Saeng becomes familiar with some aspects of her new environment, such as the geese, so “no new hope” is too absolute. Her changing attitude suggests gradual adaptation. \\
& The idea you call “representation” is not clearly explained. If you mean that the winter hibiscus represents Saeng, state this directly: like the plant, she is living in an unfamiliar, cold environment, yet she is strong enough to survive and eventually bloom. Connect this idea to her mother’s observation that the plant is “strong enough to survive the cold.” \\
& Several language errors sometimes obscure your meaning. For example, revise “the author choose” to “the author chooses,” “He memories” to “Her memories,” “drivers test” to “driver’s test,” and “tews” to “tears.” Replace “giving to how” with “showing how,” and “purly from the representation” with a precise explanation of the symbol. A clearer thesis could be: “The author ends with this paragraph to show that Saeng’s grief has become resilience and that, like the hibiscus, she can survive and grow in her new environment.” \\
\addlinespace
Qwen3.5-9B & Your response attempts to address the prompt by linking the final paragraph to themes of tone, hope, and cultural representation. You correctly identify that Saeng's emotional state shifts from despair to determination, noting her tears in the greenhouse as a pivotal moment. The detail about her being "too shaky to say the word at home" effectively captures Saeng's struggle with dual identities, supporting the idea that the story moves from loss toward adaptation. However, your explanation of *why* the author chose this specific ending relies heavily on restating Saeng's future plans rather than analyzing the underlying symbolism or character growth that makes this moment powerful. \\
& To strengthen your argument, you must move beyond summarizing Saeng's vow and explain how it resolves the story's tension. The author does not simply end the story; ends it with the *condition* of spring to symbolize Saeng's internal readiness. Your mention of the geese becoming familiar is a strong starting point, but you can deepen the analysis by connecting it to the grandmother's comment about the winter hibiscus being "strong enough to survive the cold." This metaphor directly parallels Saeng's own resilience; just as the flower thrives in harsh conditions, she is now prepared to face her failure and try again. By focusing on this parallel, you explain the author's choice more profoundly than by simply stating she will take the test. \\
& Furthermore, your writing contains several grammatical and mechanical issues that obscure your sophisticated ideas. For instance, the phrase "depress ful mood" should be corrected to "depressing mood" or "despondent mood" to maintain professional tone. Similarly, "emence emotion's" must be revised to "intense emotions," and "sr" likely means "since." The sentence structure often becomes run-on, such as "The winter starts to give her the feels all hope is lost," which should be split or punctuated clearly to ensure clarity. \\
& In conclusion, while your interpretation hits key emotional beats, your analysis lacks depth regarding the author's specific narrative choices. Focus on explaining how the imagery of spring and the specific plant choice resolves the conflict introduced at the beginning. Additionally, proofread carefully to fix spelling and punctuation errors so your powerful insights are not lost to unclear phrasing. Strengthening these areas will significantly improve the quality of your essay. \\
\addlinespace
RLAES w/o AGFO & The essay presents a nuanced and emotionally resonant interpretation of the story's tonal arc, effectively tracing the trajectory from despair through sorrow, alienation, and emotional fragmentation to eventual hope, resilience, and forward-looking determination. The writer adeptly captures the duality of grief and hope: Saeng's initial devastation (failed driving test, tears in the greenhouse, shame, fear of judgment) is contrasted with the ultimate affirmation of hope—not naive optimism, but a mature, grounded resolve rooted in love, family support, cultural identity, and the promise of renewal symbolized by spring, blooming flowers, returning geese, and budding life. The tone shift is particularly well-handled: the author identifies the "depressful mood" early on, acknowledges the weight of displacement, cultural loss, and the pain of being torn between two worlds (Vietnam as emotional home vs. Canada as physical reality), yet culminates in a clear sense of agency and future possibility—"I will take that test again." The writer demonstrates strong voice and authenticity, using colloquial yet expressive language ("very upset," "gives her the feels," "tear her apart," "explosion of emence emotion's"), which reflects a teenage narrator's genuine emotional landscape while still maintaining analytical depth. Key elements—nostalgia, identity, family loyalty, shame, courage, adaptation, bitterness vs. sweetness (bitter melon vs. fast food), and symbolic hope—are woven together with personal connection and empathetic insight. The conclusion is not merely stated but felt: the ending is alive with anticipation, emotional healing, and quiet determination. The author understands that hope here is not the absence of pain, but the ability to endure it and move forward despite it. The writer also subtly critiques American assimilation ("strange things... becoming familiar") versus cultural preservation, showing how identity is both eroded and sustained through emotion, memory, and community. Despite the informal tone and occasional misspelling or imprecise diction ("emence," "shaky," "tews," "depressful"), the emotional intelligence and literary sensitivity shine through, revealing a student who connects deeply not just with the plot, but with the human experience behind the text—the courage to fail, the grace to be comforted, the strength to hope again, and the beautiful paradox that sometimes bitterness makes sweetness more meaningful, and sometimes sorrow makes joy more precious. \\
\addlinespace
\textbf{RLAES (ours)} & The author effectively addresses the core prompt by explaining why the story concludes with Saeng's vow to retry. The writing covers the surface-level plot, acknowledging the failed driver's test and the emotional weight of the story. The response generally follows the structure of presenting an idea and offering textual evidence, such as quoting the greenhouse scene and mentioning Saeng's struggle with language at home. \\
& However, the writing often relies on superficial or literal interpretations rather than digging into the deeper themes. The response characterizes the tone primarily as "depress ful" and focuses heavily on the sadness of failure. To elevate the analysis, the author needs to highlight the shift from despair to *hope*. The conclusion is not just about sadness; it is about resilience, growth, and the possibility of a fresh start. The story moves from the "winter hibiscus" (struggling in cold soil) to the concept of spring (budding, growth). The response mentions hope only briefly at the end; it would be stronger to discuss how the mother's comfort represents support, or how failing a test is a learning opportunity rather than a permanent defeat. \\
& The writing frequently uses surface-level details. Phrases like "she misses Vietnam" are covered, but the deeper cultural elements are missed, such as the contrast between American fast food and bitter melon (adaptation), or the specific meaning of the "winter hibiscus" as something strong enough to survive. The grammar contains several errors ("author choose," "emence emotion's"), which detracts from the professionalism. \\
& The response covers mostly expressed information (what happens in the story) rather than implied meanings (what the story signifies). For instance, the author mentions the flashbacks, but does not explain *why* the author chose to end on a note of agency (Saeng making a choice) rather than passivity (Saeng just crying). The writing would benefit from distinguishing between the "strange" things in America (which she is learning to accept) and the "familiar" things in Vietnam. \\
& To improve, the response should correct grammatical errors ("author chooses") and use more precise literary vocabulary (resilience, agency, foreshadowing, metaphor). The connection between the support and the main idea needs tightening. For example, instead of just saying "She can't say home," explain that this shows her internal conflict, and the vow to retake the test shows her growing confidence. The author covers the basics, but the analysis remains largely on the surface of the plot rather than exploring the author's deeper message about overcoming adversity. \\
\bottomrule
\end{tabular}
\end{table*}

\clearpage
\end{document}